\documentclass{article}

\usepackage[nonatbib, final]{neurips_ai4d3_2023}
\usepackage[numbers]{natbib}
\usepackage[utf8]{inputenc} 
\usepackage[T1]{fontenc}    
\usepackage{hyperref}       
\usepackage{url}            
\usepackage{booktabs}       
\usepackage{amsfonts}       
\usepackage{nicefrac}       
\usepackage{microtype}      
\usepackage{xcolor}         
\usepackage{graphicx}
\usepackage{multirow}
\usepackage[noend]{algorithm2e}
\usepackage{amsmath}
\title{CongFu: Conditional Graph Fusion for Drug Synergy Prediction}

\author{%
Oleksii Tsepa$^{1,2}\thanks{Equal contribution.}$ \quad Bohdan Naida$^{1,2,3}\footnotemark[1]$ \quad Anna Goldenberg$^{1,2,3}$ \quad Bo Wang$^{1,2,4,5,6}$ \\
$^1$Department of Computer Science, University of Toronto \\
$^2$Vector Institute for Artificial Intelligence\\
$^3$The Hospital for Sick Children\\
$^4$Peter Munk Cardiac Centre, University Health Network\\
$^5$Department of Laboratory Medicine and Pathobiology, University of Toronto\\
$^6$AI Hub, University Health Network \\
\texttt{\{oleksii.tsepa,bohdan.naida\}@mail.utoronto.ca}\\
\texttt{bowang@vectorinstitute.ai}\\
\texttt{anna.goldenberg@utoronto.ca}
}

\begin{document}

\maketitle

\begin{abstract}
Drug synergy, characterized by the amplified combined effect of multiple drugs, is critically important for optimizing therapeutic outcomes. Limited data on drug synergy, arising from the vast number of possible drug combinations and testing costs, motivate the need for predictive methods. In this work, we introduce CongFu, a novel Conditional Graph Fusion Layer, designed to predict drug synergy. CongFu employs an attention mechanism and a bottleneck to extract local graph contexts and conditionally fuse graph data within a global context. Its modular architecture enables flexible replacement of layer modules, including readouts and graph encoders, facilitating customization for diverse applications. To evaluate the performance of CongFu, we conduct comprehensive experiments on four datasets, encompassing three distinct setups for drug synergy prediction. CongFu achieves state-of-the-art results on 11 out of 12 benchmark datasets, demonstrating its ability to capture intricate patterns of drug synergy. Through ablation studies, we validate the significance of individual layer components, affirming their contributions to overall predictive performance. Finally, we propose an explainability strategy for elucidating the effect of drugs on genes. By addressing the challenge of predicting drug synergy in untested drug pairs and utilizing our proposed explainability approach, CongFu opens new avenues for optimizing drug combinations and advancing personalized medicine.
\end{abstract}

\section{Introduction}

Drug combination therapy is a widely adopted approach due to its numerous advantages. Unlike monotherapy, the effect of the treatment can be significantly amplified by using a combination of drugs \cite{combther}. Furthermore, drug combinations have the potential to reduce adverse effects \cite{decr_side}, decrease toxicity \cite{oneil}, and overcome drug resistance \cite{decr_res}. Multi-drug therapy can address complex diseases such as cancer \cite{cancer_1, cancer_2} or human immunodeficiency virus \cite{hiv}. However, certain drug combinations may lead to unfavorable or harmful outcomes \cite{side_effect_1, side_effect_2}, making it crucial to accurately predict synergistic drug pairs and potential side effects resulting from different drug interactions. 

Historically, the discovery of drug combinations has relied on clinical trials and trial-and-error methods. These approaches are not only costly and time-consuming but can also pose risks to patients \cite{paradigm, mixed_int}. Moreover, the scalability limitations of wet-lab tests restrict the screening of drug combinations \cite{unscale}. However, advancements in experimental techniques have led to the development of high-throughput drug screening (HTS) \cite{impact_hts, yeast_assay, methods_hts}, a fast and precise method that allows researchers to explore large drug combination spaces. This has resulted in a rapid increase in drug combination synergy data. Public databases like ASDCD \cite{asdcd} provide drug combination data and large HTS synergy studies covering numerous drugs and cancer cell lines \cite{oneil}. These databases provide high-quality training data for the development of computational approaches and aid in evaluating these methods for predicting novel drug combinations. However, the discrepancy between in vivo and in vitro experiments limits the effectiveness of HTS.

In recent years, the availability of large HTS datasets \cite{almanac} has spurred the development of machine learning models for drug synergy predictions \cite{mlbaselines}. Early deep learning methods, such as DeepSynergy \cite{deepsyn} and MatchMaker \cite{matchmaker}, utilize fully connected networks based on cell lines and drug features derived from Morgan fingerprints \cite{morgan}. Subsequent models like AuDNNsynergy \cite{audnn} incorporate autoencoders that leverage "copy number variation" data, gene expressions, and mutations. Other models like TranSynergy \cite{transynergy} adopt a transformer architecture to process a cell line and two drug feature vectors as input. DTF \cite{dtf} integrates a tensor factorization and a deep neural network for drug synergy prediction. Models like DeepDDS \cite{deepdds} and DDoS \cite{ddos} utilize graph neural networks over the molecular graph to enrich drug encoding. Further, Jiang's \cite{jiangs} and  Hu's methods \cite{pretrain} expand the range of modalities employed for drug synergy predictions, including drug-drug and drug-target interactions. SDCNet \cite{sdcnet} introduces the concept of cell line-specific graph representations for drug synergy data and trains a relational graph convolutional network over it. 

Considering that drugs interact in the context of cell line treatment, we formulate the problem as a conditional variation of drug pair scoring framework \cite{rozemberczki2021unified} and call the cell line as a context of drug interaction. Further, we refer to it as "context" for simplification. 

While existing approaches have demonstrated that sharing information between multiple modalities (fusion) leads to a performance gain \cite{bottlneck}, the fusion strategy in drug synergy is mostly a simple concatenation of latent representations, failing to capture the intrinsic dynamic synergies between drug pairs and cell lines.

Therefore, inspired by the concept of information fusion and the incorporation of a larger amount of contextual information in graph encoding, we introduce CongFu (Conditional Graph Fusion) for conditional drug pair scoring with a specific application of drug synergy prediction. The proposed layer includes context propagation and bottleneck, which work together to efficiently fuse two molecular graphs and a cell line. We present a technique for utilizing the proposed layer and evaluate the framework’s performance on 12 benchmarks. The results indicate that our architecture outperforms existing approaches, and the inclusion of the CongFu layer tends to benefit other graph-based architectures. Additionally, we conduct ablation studies to emphasize the importance of every component in the proposed layer. Our explainability framework helps interpret model predictions, revealing the impact of drugs on specific genes.

To sum up, our contribution can be stated as follows:
1) We propose a novel CongFu layer for conditional graph pair scoring and apply it to drug synergy predictions
2) We conduct an ablation study to highlight the importance of fusion between graphs and to explore an appropriate place for initiating information sharing
3) We set the new state-of-the-art for 11 benchmarks derived from the DrugComb database in inductive and transductive setups
4) We provide the interpretability of our model to gain biological insights on gene-drug interactions.

\section{Related Work}

The related works can be categorized as follows:

\textbf{Linear Models.} Models such as Deep Synergy \cite{deepsyn}, MatchMaker \cite{matchmaker}, and AuDNNsynergy \cite{audnn} utilize fully connected networks to process cell lines and drug features encoded via Morgan fingerprints. Deep Synergy applies a single MLP over the concatenated input triplet, while MatchMaker uses one MLP with shared weights to encode each drug conditionally based on the cell line. The hidden representations of the drug pairs are then passed to the MLP. AuDNNsynergy has a similar architecture to Deep Synergy but additionally processes "copy number variation" data, gene expressions, and mutations via autoencoders.

\textbf{Graph-based methods.} Models like DeepDDS \cite{deepdds} and DDoS \cite{ddos} employ Message Passing Neural Networks (MPNNs) to encode each graph separately and an MLP to encode the cell line. All processed modalities are then concatenated and passed to the MLP. In models like SDCNet \cite{sdcnet} and Jiang's method \cite{jiangs}, the problem of drug synergy is formulated as link prediction. Both methods create cell line-specific heterogeneous networks of drugs and utilize an encoder-decoder architecture. Additionally, Jiang's method incorporates proteins into the drug-drug network, while Hu et al. \cite{pretrain} construct a single heterogeneous network of cell lines, drugs, diseases, and proteins. They use the RotatE model \cite{rotate} to encode diseases and pre-trained models \cite{kpgt, esm1b} for other modalities of encoding. After propagation in the heterogeneous graph, drug and cell line embeddings are passed to the MLP for the final prediction.

\textbf{Our method: CongFu.} Our proposed method, CongFu, is a significant advancement in the graph-based category. We introduce a novel layer, CongFu, which formulates the heterogeneous graphs between multiple drugs. This layer models dynamic interactions in a nonlinear manner, representing a substantial improvement over existing methods. Additionally, we investigate a variety of strategies to optimally integrate the CongFu layer, with the aim of maximizing predictive performance.

\section{Methods}

\begin{figure}
    \centering
    \includegraphics[width=1\textwidth]{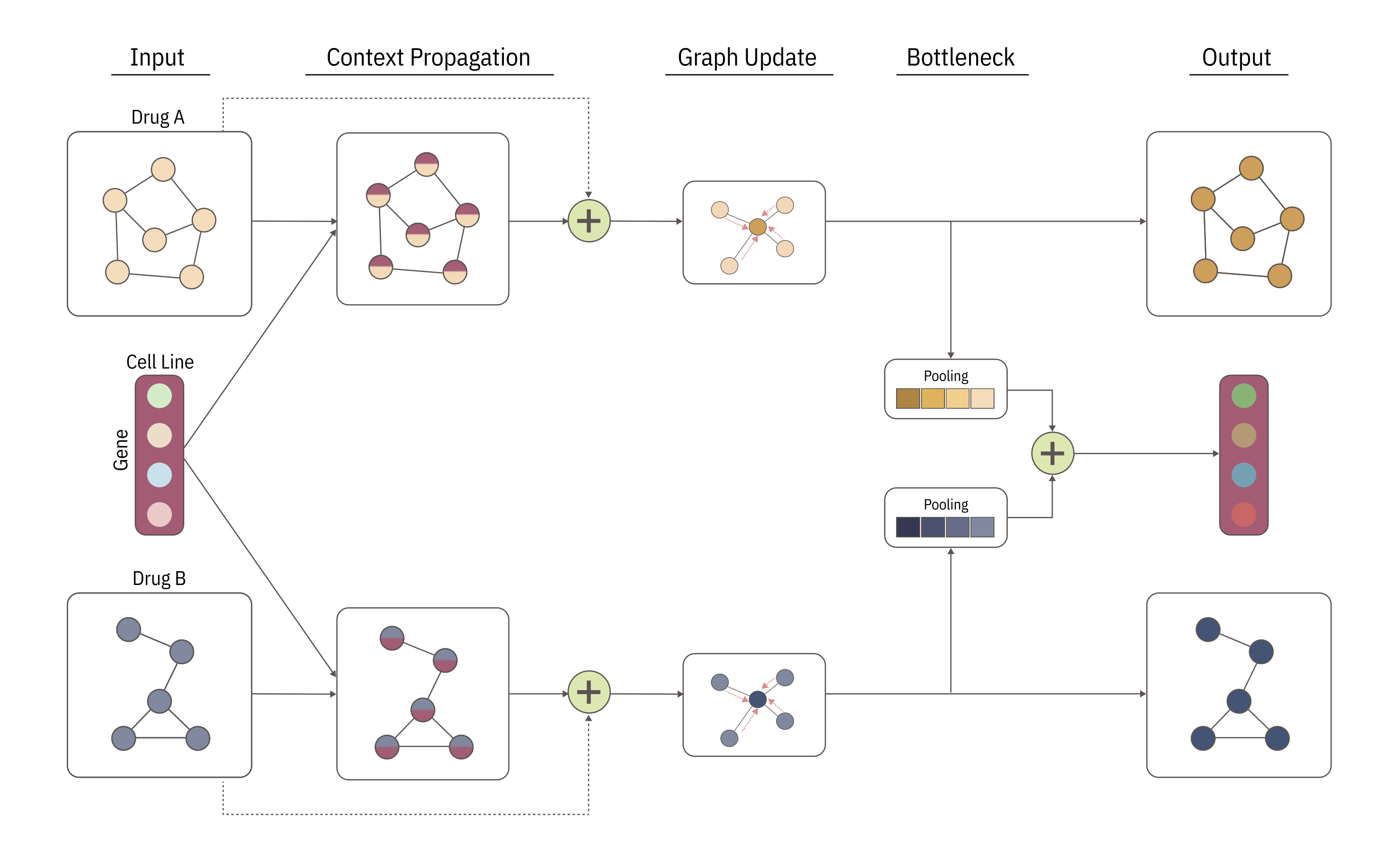}
    \caption{CongFu layer architecture. The layer takes in two graphs and a context vector as its input and produces an updated triplet as the output. This layer consists of three separate modules: Context Propagation, which facilitates the transfer of information from the initial context to the nodes of the graphs; Graph Update, which propagates the injected information throughout the graphs; and a Bottleneck that merges the local contexts related to each graph into a global context. The resulting output comprises updated representations of the two graphs and the context.}
    \label{congfu-figure}
\end{figure}

This section formalizes the CongFu layer and its components (Fig. \ref{congfu-figure}). We start with the problem formulation including notations and task descriptions. We then break down the architecture of the CongFu layer into corresponding equations. Finally, we discuss its modularity characteristic, explore potential use cases, and consider possible adaptation of the algorithm to support problems with multiple (exceeding $2$) graphs.

\subsection{Problem Formulation}

\textbf{Notations}.
Let $\mathcal{G} = (\mathcal{V}, \mathcal{E})$ denote a graph with $N$ nodes and $E$ edges. Graph $\mathcal{G}$ is associated with an adjacency matrix $\mathbf{A} \in \mathbb{R}^{N \times N}$, where $\mathbf{A}_{ij} = 0$ if there is no edge between nodes $i$ and $j$; node feature set is denoted as $\mathbf{X} \in \mathbb{R}^{N \times D_{node}}$, while edge feature set is denoted as $\mathbf{E} \in \mathbb{R}^{E \times D_{edge}}$

\textbf{Task: Conditional drug pair scoring.} Given a set of graph pairs $\mathcal{D} = \{(\mathcal{G}_{A}^{1}, \mathcal{G}_{B}^{1}),..., (\mathcal{G}_{A}^{n}, \mathcal{G}_{B}^{n})\}$ and associated context features for each graph pair $\mathcal{C} = \{\mathbf{C}_1, …, \mathbf{C}_n\}$ as input, where $\mathbf{C}_i \in \mathbb{R}^{D_{cont}}$, the objective of the task is to predict the corresponding target values $\mathbb{Y} = \{\mathbf{Y}_1, …, \mathbf{Y}_n\}$, where $\mathbf{Y_i}$ is a scalar value.

\subsection{CongFu Layer Architecture}

The CongFu layer receives as input two molecular graphs and a context vector (representing a cell line) and outputs an updated triplet. This layer consists of three distinct modules:

\begin{enumerate}
    \item Context Propagation: allows the flow of information from the initial context to the nodes of the graphs
    \item Graph Update: propagates the injected information along the graph
    \item Bottleneck: combines the local contexts associated with each graph to form a global context
\end{enumerate}
As a result, the layer outputs updated representations of both graphs and the context.

The context features $\mathbf{C} \in \mathbb{R}^{1 \times D_{cont}}$ are required to have the same dimension $(D_{cont})$ as node features $\mathbf{X}_A \in \mathbb{R}^{N_A \times D_{node}}, \mathbf{X}_B \in \mathbb{R}^{N_A \times D_{node}} $ $(D_{node})$. To achieve this, the context features are linearly transformed as illustrated in (\ref{linear_transform}), where $\mathbf{W} \in \mathbb{R}^{D_{cont} \times D_{node}}$. If context and node feature dimensions are equal $(D_{cont} = D_{node})$, no transformation is required. Further, node and context feature dimensions are denoted as $D$. 

\begin{equation}\label{linear_transform}
\mathbf{C} = \mathbf{CW}
\end{equation}

The modular structure of the CongFu layer allows for the free choice of an aggregation strategy of initial and updated feature sets in (\ref{propagation}), an MPNN for the \textit{Graph Update} module (\ref{graph_update_mpnn}), and a replacement of readout in the Bottleneck module (\ref{F_bottle}).

\subsubsection{Context propagation}
This module updates node feature representations $\mathbf{X}_A$ and $\mathbf{X}_B$ based on the context $\mathbf{C}$ using a conditional approach. The updated node representations $\hat{\mathbf{X}}_A, \hat{\mathbf{X}}_B$ are then added to the initial node representations $\mathbf{X}_A, \mathbf{X}_B$ (\ref{propagation}). Specifically, we can express this process as follows:

\begin{equation}\label{propagation}
\hat{\mathbf{X}}_{j} = \mathbf{X}_{j} + \mathbf{W}_1 \mathbf{X}_{j} + \mathbf{W}_2 \texttt{ReLU}(\mathbf{C} \mathbf{W}_3)
\end{equation}

Here, $j \in \{A, B\}$, $\{\mathbf{X}_A, \mathbf{X}_B, \hat{\mathbf{X}}_A, \hat{\mathbf{X}}_B\} \in \mathbb{R}^{N \times D}$, while $\{\mathbf{W}_1, \mathbf{W}_2, \mathbf{W}_3\} \in \mathbb{R}^{D \times D}$.

\subsubsection{Graph update}
After injecting the context into the node features, information is propagated along the graph.
Due to the module's modularity property, any MPNN (e.g., GIN \cite{gin}, GraphSAGE \cite{sage}, GPS \cite{gps}) can be used for graph updates (\ref{graph_update_mpnn}). The edge features $\mathbf{E}$ are optional and can be passed if available, and the MPNN supports them. Then, Batch Normalization \cite{bn} and ReLU \cite{relu} are applied over $\hat{\mathbf{X}}_{j}$, where $j \in {A, B}$.

\begin{equation} \label{graph_update_mpnn}
\hat{\mathbf{X}}_{j} = \texttt{MPNN}(\hat{\mathbf{X}}_{j}, \mathbf{E}_j, \mathbf{A}_{j})
\end{equation}

\begin{equation} \label{graph_update_bn}
\hat{\mathbf{X}}_{j} = \texttt{BatchNorm}(\hat{\mathbf{X}}_{j})
\end{equation}

\begin{equation} \label{graph_update_relu}
\hat{\mathbf{X}}_{j} = \texttt{ReLU}(\hat{\mathbf{X}}_{j})
\end{equation}

\subsubsection{Bottleneck}

In this step, the attention-based readout function $\mathcal{F}$ is used to aggregate node-level features $\hat{\mathbf{X}}_A, \hat{\mathbf{X}}_B$ conditioned on the context $\mathbf{C}$ (\ref{F_bottle}). The resulting output, $\mathbf{C}_A, \mathbf{C}_B$, represents the local context of each graph.

\begin{equation}\label{F_bottle}
\mathcal{F}(\hat{\mathbf{X}}_{j}; \mathbf{C}) \rightarrow \mathbf{C}_{j}
\end{equation}

The equation of the readout function $\mathcal{F}$, inspired by the Graph Attention Networks (GAT) \cite{gat} layer update rule, is illustrated in (\ref{update_gat}), where $j \in {A, B}$, $\mathbf{W} \in \mathbb{R}^{D \times D}$, ${d,b} \in mathbb{R}^{D}$ are trainable matrix and biases respectively. The attention coefficient $a_i$ is computed in (\ref{att_score_gat}) to account for the importance of context and node features.

\begin{equation}\label{update_gat}
\mathbf{C}_{j} =   \sum_{i}^{N_{j}}  \frac{1}{N_{j} + 1} \alpha_{i} \mathbf{W} \mathbf{X}_{j}^{i} + b
\end{equation}

\begin{equation}\label{att_score_gat}
\alpha_{i} = \texttt{softmax}(d^{T} [\mathbf{W} \mathbf{C} || \mathbf{W} \mathbf{X}_{j}^{i}])
\end{equation}

Finally, we update the global context $\hat{\mathbf{C}}$ as a sum of local contexts $\mathbf{C}_{A}$ and $\mathbf{C}_{B}$ (\ref{global_fuse_module}).

\begin{equation}\label{global_fuse_module}
\hat{\mathbf{C}} = \mathbf{C}_A + \mathbf{C}_B
\end{equation}

\subsubsection{Extending to multiple graphs}
While our algorithm is designed to operate over two graphs, it can be easily extended for multiple (more than $2$) graphs. The \textit{Context Propagation} and \textit{Graph Update} modules are invariant to the number of graphs as each receives a single graph as input. The fusion process remains consistent within the \textit{Bottleneck}. Consequently, the sum over two graphs in (\ref{global_fuse_module}) is replaced by the sum over $\textit{k}$ graphs' contexts.

\begin{equation}\label{sum_fusion}
\hat{\mathbf{C}} = \sum_{j}^{k} \mathbf{C}_j 
\end{equation}

\section{Experiments}

\subsection{Datasets}

We perform an extensive evaluation of the proposed algorithm for predicting drug synergy effects using four datasets tested in three different setups, resulting in a total of 12 benchmarks. These datasets are sourced from DrugComb \cite{drugcomb}, the most comprehensive and current database of drug combinations. The goal is to predict the synergistic effect of drugs for a specific cell line (referred to as the context in our notation), with the drugs represented as SMILES \cite{weininger1988smiles}. More details on the dataset can be found in Appendix \ref{app:dataset}.

\subsection{Experimental Setup}

We assess the model using three distinct setups for each dataset, adhering to the methodology proposed by DeepDDS \cite{deepdds}. Specifically, we employ a transductive setup with a 5-fold cross-validation, where the training set is further divided into training and validation subsets using a 90/10 ratio. In the leave-drug-out setup, we partition the set of drugs into five equally sized groups, with the training set excluding all drugs from the test set. We then conduct cross-validation stratified by drug groups. For the leave-combination-out setup, the drug pairs from the test set are removed from the training set, although individual drugs may still appear in both the training and test sets. The performance of the model is evaluated using AUROC and AUPRC metrics to deal with imbalance.

\subsection{Implementation details}

The atom representation is computed as the embedded atomic number, while the edge representation corresponds to the embedded bond type. The cell line is compressed into the latent space via a $2$-layer MLP. Each graph is individually encoded using a $3$-layer Graph Isomorphism Network (GINE) \cite{gine} with Batch Normalization \cite{bn} and ReLU activation. Two CongFu layers with the same embedding dimension are applied to integrate information from the graphs and the cell line, utilizing GINE as an MPNN. A prediction head consisting of a $2$-layer MLP operates over the concatenation of the cell line and drug representations. ReLU is used as an activation function between hidden layers in all MLPs. The training was conducted on a single NVIDIA RTX $6000$ taking approximately $2$ minutes per epoch. We utilized the Adam \cite{adam} optimizer and binary cross-entropy loss during training. The training setup and hyperparameters remained consistent across all benchmarks, as illustrated in Appendix \ref{app:hyperparams}. The overall architecture is depicted in Appendix \ref{app:schematics}.

\begin{table}
  \caption{Comparison to SOTA on DrugComb - HSA Synergy Score}
  \label{hsa_table}
  \centering
  \resizebox{\textwidth}{!}{
  \begin{tabular}{lcccccc}
    \toprule
    \multirow{2}{*}{Method} & \multicolumn{2}{c}{Transductive} & \multicolumn{2}{c}{Leave-comb-out} & \multicolumn{2}{c}{Leave-drug-out}\\
    \cmidrule(l){2-7}
    & AUROC & AUPRC & AUROC & AUPRC & AUROC & AUPRC \\
    \midrule
    CongFu (ours) & $\mathbf{0.976 \pm 0.001}$ & $\mathbf{0.949 \pm 0.002}$ & $\mathbf{0.968 \pm 0.003}$ & $\mathbf{0.931 \pm 0.007}$ & $\mathbf{0.832 \pm 0.02}$ & $\mathbf{0.67 \pm 0.03}$ \\
    DeepDDS & $0.956 \pm 0.02$ & $0.913 \pm 0.005$ & $0.937 \pm 0.008$ & $0.87 \pm 0.018$ & $0.798 \pm 0.03$ & $0.625 \pm 0.04$ \\
    DSN - DDI &  $0.931 \pm 0.01$ & $0.861 \pm 0.02$ & $0.948 \pm 0.004$ & $0.894 \pm 0.006$ & $0.8 \pm 0.013$ & $0.645 \pm 0.031$ \\
    XGBoost & $0.73 \pm 0.005$ & $0.565 \pm 0.008$ & $0.729 \pm 0.005$ & $0.556 \pm 0.003$ & $0.684 \pm 0.022$ & $0.48 \pm 0.045$ \\
    LogReg & $0.723 \pm 0.005$ & $0.536 \pm 0.007$ & $0.718 \pm 0.006$ & $0.528 \pm 0.013$ & $0.67 \pm 0.018$ & $0.435 \pm 0.043$ \\
    \bottomrule
  \end{tabular}
  }
\end{table}

\begin{table}
  \caption{Ablation study on a conditional fusion on DrugComb - HSA Synergy Score}
  \label{sample-table}
  \centering
  \resizebox{\textwidth}{!}{
  \begin{tabular}{lcccccc}
    \toprule
    \multirow{2}{*}{Method} & \multicolumn{2}{c}{Transductive} & \multicolumn{2}{c}{Leave-comb-out} & \multicolumn{2}{c}{Leave-drug-out}\\
    \cmidrule(l){2-7}
    & AUROC & AUPRC & AUROC & AUPRC & AUROC & AUPRC \\
    \midrule
    CongFu (ours) & $\mathbf{0.976 \pm 0.001}$ & $\mathbf{0.949 \pm 0.002}$ & $\mathbf{0.968 \pm 0.003}$ & $\mathbf{0.931 \pm 0.007}$ & $\mathbf{0.832 \pm 0.02}$ & $\mathbf{0.67 \pm 0.03}$ \\
    
    w/o conditioning & $0.966 \pm 0.001$ & $0.928 \pm 0.003$ & $0.949 \pm 0.002$ & $0.892 \pm 0.007$ & $0.812 \pm 0.018$ & $0.633 \pm 0.04$ \\
    
    w/o fusion &  $0.97 \pm 0.017$ & $0.939 \pm 0.03$ & $0.955 \pm 0.002$ & $0.906 \pm 0.006$ & $0.81 \pm 0.02$ & $0.607 \pm 0.04$ \\
    \bottomrule
  \end{tabular}
  }
\end{table}

\subsection{Results}

Our model, based on the CongFu layer, demonstrates superior performance compared to the other methods in 11 out of the 12 benchmarks, according to the AUPRC score. The only exception is observed in the Loewe leave-combination-out setup, where DeepDDS \cite{deepdds} marginally surpasses our model by $0.003$ in terms of AUROC. However, it's worth noting that our model still outperforms DeepDDS by $0.02$ in AUPRC for the same setup. The most significant performance gap in favor of our model is observed in the benchmarks using the HSA score (Table \ref{hsa_table}), with an improvement ranging from $0.036$ to $0.061$ in AUPRC. Due to the large input vector of size $1508$, XGBoost and Logistic Regression struggle to capture all dependencies and show poor performance compared to the state-of-the-art models. Tables \ref{hsa_table}, and other quantitative results in Appendix \ref{sec:add_experiments}, summarize our results and compare CongFu-based architecture to other models mentioned in the experimental setup section. Importantly, our model exhibits substantial improvement in inductive settings, underscoring its capacity to generalize effectively to unseen data.

\subsubsection{Ablation study}

In order to validate the significance of conditional fusion, we perform an ablation study, focusing on the transductive and leave-drug-out HSA benchmarks. The study is comprised of two main experiments:

\begin{itemize}
\item \textit{Without conditioning}: The aim of this experiment is to assess the impact of conditioning on the context in the fusion component. We substitute the \textit{Context Propagation} module in CongFu with a cross-attention module. In this module, all nodes of $\mathcal{G}_A$ and $\mathcal{G}_B$ are interconnected through a bipartite graph, and information exchange is facilitated via a GAT as implemented in the intra-view of DSN-DDI. While this approach enables information sharing between the two graphs, it does not account for the context between them.

\item \textit{Without fusion}: This experiment is designed to evaluate the importance of information sharing (fusion) between graphs. We replace all CongFu layers with MPNNs - specifically, GINE - to encode drugs independently. The results of this ablation study highlight the significance of information exchange between the two graphs.
\end{itemize}

\subsubsection{Determining the Optimal Point for Fusion}

We conduct a series of experiments using a 5-layer model composed of MPNNs and fusion layers (either CongFu or cross-attention layers). The fusion layers are applied after the MPNNs, which initially encode each graph independently, without any information exchange. The idea behind the middle fusion is, firstly, to let the model learn representations of drugs separately, then learn the interaction of each drug with the cell line (as it was in MatchMaker), and finally combine representations of drugs and a cell line together. The aim of this study is to highlight the importance of conditional fusion (CongFu) between graphs and to identify the optimal point for initiating information sharing, referred to as 'fusion injection'.

\begin{figure}[h]
\centering
    \includegraphics[width=1\textwidth]{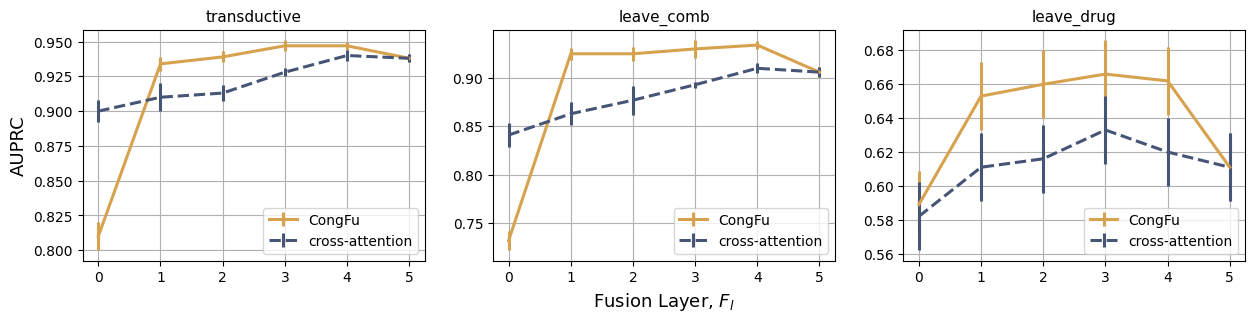}
    \caption{The impact of using the conditional fusion starting from different model layers on the leave-drug-out HSA benchmark}
    \label{fusion-inject}
\end{figure}

On Figure \ref{fusion-inject}, the x-axis (Fusion layer) represents the layer in the model where fusion begins, i.e., the number of preceding MPNN layers. For instance, for $F_l = 2$, the model consists of $2$ MPNN layers followed by $3$ fusion layers. Overall, starting from $F_l = 1$, the models with CongFu layers consistently outperform those with cross-attention layers. The study reveals that in transductive and leave-drug-out setups, the optimal model configuration includes $3$ MPNN layers and $2$ CongFu layers. However, for the leave-comb setup, the model requires $4$ MPNNs.

\subsubsection{Explainability}

In this chapter, we provide biological insights by elucidating the predictions of the model. Our aim is to answer the question: “What impact does each drug have on a specific gene?” To tackle this, we use the chain rule to estimate the gradient magnitude of the output passed through drug encoders w.r.t a specific gene. 

Firstly, we calculate gradients of pooled drug embeddings from the last CongFu layer w.r.t a gene. The gene is represented as $g \in \mathbb{R}$, the cell line is denoted as $C = [g_1, g_2 ... g_n] \in \mathbb{R}^N$, and drug embedding is denoted as $h \in \mathbb{R}^D$.
\begin{equation}\label{explain_1}
\mathcal{R}(g_i, h) = \left[\mathcal{R}(g_i, h_1), \mathcal{R}(g_i, h_2), ..., \mathcal{R}(g_i, h_d) \right] = \left[\frac{\partial h_1}{\partial g_i}, \frac{\partial h_2}{\partial g_i}, ..., \frac{\partial h_n}{\partial g_i} \right]
\end{equation}

Next, we compute gradients of the model output from the predictive head,  denoted as $y$, w.r.t the drug embeddings obtained from the last CongFu layer.
\begin{equation}\label{explain_2}
\mathcal{R}(h, y) = \left[ \mathcal{R}(h_1, y), \mathcal{R}(h_2, y), ..., \mathcal{R}(h_d, y) \right] = \left[ \frac{\partial y}{\partial h_1}, \frac{\partial y}{\partial h_2}, ..., \frac{\partial y}{\partial h_d}\right]
\end{equation}

Finally, we calculate the modulus of the dot product between these gradients multiplied by the input value (gene), which represents the magnitude of the gradient passed through the drug encoder.
\begin{equation}\label{explain_3}
\mathcal{R}(g_i) = \left| g_i \mathcal{R}(g_i, h) \mathcal{R}(h, y) \right|= \left| g_i \sum_j \frac{\partial h_j}{\partial g_i} \frac{\partial y}{\partial h_j} \right|  
\end{equation}

To assess the impact of each drug on a gene, we compute the proportion of the magnitudes of drug A and drug B.

Prior research \cite{explain_1, explain_2, explain_3} has demonstrated that the combination of the epidermal growth factor receptor (EGFR) inhibitor Afatinib and the serine/threonine protein kinase B (AKT) inhibitor MK2206 has a synergistic impact on the treatment of lung cancer and head and neck squamous cell carcinoma. In DeepDDS \cite{deepdds}, authors plot heat maps of the atom correlation matrix before and after training to observe the change in feature patterns.

\begin{figure}[h]
    \centering
    \includegraphics[width=0.6\textwidth]{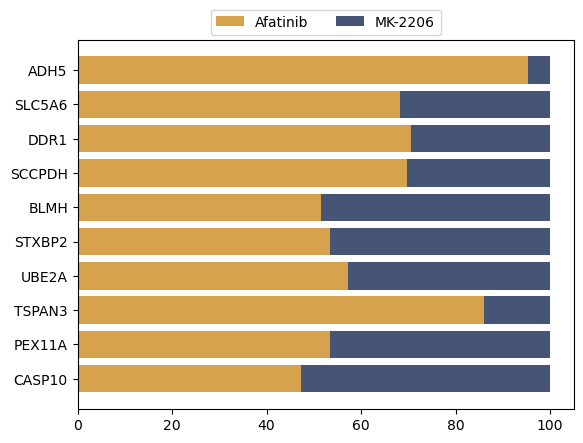}
    \caption{Impact of Afatinib and MK2206 top most important genes from BT-20 cell line}
    \label{explain-figure}
\end{figure}

Figure \ref{explain-figure} illustrates the impact of Afatinib and MK2206 on the top 10 most important genes from BT-20 (breast tumor) cell line, sorted in descending order according to the magnitude of gradients. Interestingly, these drugs have different impacts on each gene. For example, genes ADH5 and TSPAN3 primarily interact with Afatinib rather than MK-2206. Though both drugs have quite similar impacts on UBE2A and PEX11A genes. We believe the provided explainability framework will offer scientists interesting biological insights regarding model prediction and will expedite the discovery of new drugs.

\section{Conclusion}
In this work, we introduced a novel Conditional Graph Fusion Layer (CongFu) specifically designed for drug synergy predictions. The CongFu layer utilizes an attention readout mechanism and a bottleneck module to extract local graph contexts and conditionally fuse graph data within a global context. The modular design of CongFu allows for easy customization by replacing layer modules, such as readouts and graph encoders.

We conducted extensive experiments on four distinct datasets across three different setups to evaluate CongFu's performance in predicting drug synergy. CongFu outperformed state-of-the-art methods on $11$ out of $12$ benchmark datasets, demonstrating its ability to capture complex drug synergy patterns. Ablation studies further confirmed the importance of incorporating CongFu layers and their contribution to the overall predictive performance.

By effectively predicting drug synergy in untested drug pairs, CongFu paves the way for optimizing drug combinations and advancing personalized medicine. However, our study is not without limitations. While we have developed a universal technique for solving the conditional graph pair scoring problem, it is currently only applicable in the domain of drug synergy prediction. As new problems that align with the task's requirements emerge, our methodology can be applied and tested on them. Although it is theoretically possible to extend the approach to a broader range of input graphs, the lack of appropriate datasets prevents us from evaluating CongFu's performance on these problems. To the best of our knowledge, language models (LMs) have not been applied yet for predicting drug synergy. We will consider combining LMs with the idea of fusion in future work.



\newpage

{
\small
\bibliographystyle{unsrt}
\bibliography{bibtex}
}


\appendix
\renewcommand\thefigure{\thesection.\arabic{figure}}
\renewcommand\thetable{\thesection.\arabic{table}}

\section{Dataset details}\label{app:dataset}
Drugs represented as SMILES were converted using RDKit \cite{rdkit} into a PyG \cite{pyg} graph, with atoms represented as nodes and bonds represented as edges. The features of the cell line are gathered from the Genomics of Drug Sensitivity in Cancer \footnote{https://www.cancerrxgene.org/gdsc1000/GDSC1000\_WebResources/Data/preprocessed/\\/Cell\_line\_RMA\_proc\_basalExp.txt.zip}. This includes normalized basal expression profiles of approximately 1000 human cancer cell lines. From the normalized expression levels of 17737 genes, we select 908 landmark genes \cite{l1000}. Therefore, the cell lines are represented by a feature vector of length 908.

\section{Baselines}\label{app:baselines}
CongFu is benchmarked against state-of-the-art methods for drug synergy predictions. Specifically, we utilize the official implementations of DeepDDS \cite{deepdds} and a modified version of DSN-DDI \cite{dsnddi}. In the latter, the relation-type embedding is substituted with a cell feature matrix, enabling the prediction of drug synergy from drug-drug interactions. For the baseline models, we implement Logistic Regression \cite{logreg} and XGBoost \cite{xgboost} over the concatenated representations of cell lines and drugs, encoded using a pre-trained Deep Graph Infomax model \cite{gine}. SDCNet \cite{sdcnet} and Hu's methods \cite{pretrain} are not included in the comparison due to the irreproducibility and absence of codebase, respectively.

\section{Preprocessing description} \label{app:preprocessing}

Each dataset is created by quantifying the target through four distinct types of synergy scores, specifically Loewe additivity (Loewe) \cite{loewe}, Bliss independence (Bliss) \cite{bliss}, zero interaction potency (ZIP) \cite{zip}, and highest single agent (HSA) \cite{hsa}. These targets describe the measurement of drug interaction, specifically the degree of additional drug responses observed compared to the expected response. In other words, drug synergy indicates the percentage of excess or reduced response in antagonistic settings.

The preprocessing of the dataset follows the strategy of DDoS \cite{ddos}. Initially, we exclude all triplets that do not have corresponding identifiers in the cell line feature table. Then, triplets with any missing data (cell line, drugs, targets) are filtered out. Finally, duplicated triplets are removed. Each of the four synergy scores is binarized based on thresholds \cite{methods_hts}. Samples with a synergy score above 10 are considered positive (synergistic), and samples lower than -10 are considered negative (antagonistic). Consequently, we end up with four datasets (Loewe, Bliss, HSA, and ZIP) with names corresponding to their targets. The statistics of each dataset are described in Table \ref{benchmark-datasets}.

\begin{table}[h]
  \caption{Statistics of the datasets, where Loewe, Bliss, HSA, ZIP - datasets derived from DrugComb. Statistics for DrugComb are calculated after the preprocessing stage. The percentage of the positive labels is rounded to the first decimal point.}
  \label{benchmark-datasets}
  \centering
  \resizebox{0.7\textwidth}{!}{
  \begin{tabular}{lllll}
    \toprule
    \cmidrule(r){1-2}
    Dataset & $\#$ Samples & $\%$ Positive Labels & $\#$ Drugs & $\#$ Cell Lines \\
    \midrule
    Loewe    & 163816 & 14.8 & 2147 & 164 \\
    Bliss    & 125548 & 49.5 & 1868 & 164 \\
    HSA      & 108559 & 29.5 & 1189 & 162 \\
    ZIP      & 89047  & 59.8 & 1810 & 162 \\
    \bottomrule
    DrugComb  & 647232 & - & 4268 & 288
  \end{tabular}
  }
\end{table}

\newpage

\section{Model hyperparameters}\label{app:hyperparams}

\begin{table}[h]
  \caption{Model hyperparameters}
  \label{hyperparams-table}
  \centering
  \resizebox{12cm}{!}{
  \begin{tabular}{ll|ll}
    \toprule
    Hyperparameter     & Value & Hyperparameter     & Value \\
    \midrule
    Learning rate & $1e-4$ & Epochs & $100$\\
    Node embedding size & $300$ & Edge embedding size & $300$ \\
    $\#$ Graph encoders & $3$ & Graph encoder  & $\texttt{GINE}([300,300,300])$   \\
    $\#$ Graph encoders (CongFu)  & $2$ &  Graph encoder (CongFu)  & $\texttt{GINE}([300,300,300])$ \\
    Cell line encoder & $[908,512,300]$ & Prediction Head & $[812,256,64]$  \\
    \bottomrule
  \end{tabular}
  }
\end{table}

\section{Additional experimental results}\label{sec:add_experiments}

\begin{table}[h]
  \caption{Comparison to SOTA on DrugComb - Bliss Synergy Score}
  \label{bliss_table}
  \centering
  \resizebox{\textwidth}{!}{
  \begin{tabular}{lcccccc}
    \toprule
    \multirow{2}{*}{Method} & \multicolumn{2}{c}{Transductive} & \multicolumn{2}{c}{Leave-comb-out} & \multicolumn{2}{c}{Leave-drug-out}\\
    \cmidrule(l){2-7}
    & AUROC & AUPRC & AUROC & AUPRC & AUROC & AUPRC \\
    \midrule
    CongFu (ours) & $\mathbf{0.982 \pm 0.001}$ & $\mathbf{0.981 \pm 0.001}$ & $\mathbf{0.975 \pm 0.002}$ & $\mathbf{0.974 \pm 0.003}$ & $\mathbf{0.79 \pm 0.02}$ & $\mathbf{0.779 \pm 0.02}$ \\
    DeepDDS & $0.956 \pm 0.004$ & $0.955 \pm 0.004$ & $0.941 \pm 0.009$ & $0.938 \pm 0.009$ & $0.76 \pm 0.03$ & $0.75 \pm 0.03$ \\
    DSN - DDI &  $0.894 \pm 0.04$ & $0.886 \pm 0.04$ & $0.952 \pm 0.003$ & $0.946 \pm 0.004$ & $0.754 \pm 0.005$ & $0.742 \pm 0.009$ \\
    XGBoost & $0.717 \pm 0.003$ & $0.652 \pm 0.004$ & $0.712 \pm 0.005$ & $0.647 \pm 0.011$ & $0.64 \pm 0.009$ & $0.584 \pm 0.012$ \\
    LogReg & $0.664 \pm 0.003$ & $0.605 \pm 0.003$ & $0.661 \pm 0.004$ & $0.603 \pm 0.01$ & $0.595 \pm 0.011$ & $0.552 \pm 0.012$ \\
    \bottomrule
  \end{tabular}
  }
\end{table}

\begin{table}[h]
  \caption{Comparison to SOTA on DrugComb - Loewe Synergy Score}
  \label{loewe_table}
  \centering
  \resizebox{\textwidth}{!}{
  \begin{tabular}{lcccccc}
    \toprule
    \multirow{2}{*}{Method} & \multicolumn{2}{c}{Transductive} & \multicolumn{2}{c}{Leave-comb-out} & \multicolumn{2}{c}{Leave-drug-out}\\
    \cmidrule(l){2-7}
    & AUROC & AUPRC & AUROC & AUPRC & AUROC & AUPRC \\
    \midrule
    CongFu (ours) & $\mathbf{0.939 \pm 0.005}$ & $\mathbf{0.791 \pm 0.01}$ & $0.772 \pm 0.04$ & $0.403 \pm 0.07 $ & $0.774 \pm 0.03$ & $\mathbf{0.423 \pm 0.07}$\\
    DeepDDS & $0.926 \pm 0.004$ & $0.746 \pm 0.018$ & $\mathbf{0.775 \pm 0.03}$ & $\mathbf{0.409 \pm 0.06}$ & $\mathbf{0.777 \pm 0.027}$ & $0.403 \pm 0.07$ \\
    DSN - DDI &  $0.807 \pm 0.02$ & $0.437 \pm 0.039$ & $0.771 \pm 0.028$ & $0.358 \pm 0.02$ & $0.774 \pm 0.019$ & $0.361 \pm 0.033$ \\
    XGBoost & $0.621 \pm 0.002$ & $0.303 \pm 0.003$ & $0.562 \pm 0.013$ & $0.215 \pm 0.026$ & $0.562 \pm 0.013$ & $0.215 \pm 0.026$ \\
    LogReg & $0.61 \pm 0.004$ & $0.27 \pm 0.005$ & $0.58 \pm 0.02$ & $0.21 \pm 0.028$ & $0.58 \pm 0.02$ & $0.21 \pm 0.028$ \\
    \bottomrule
  \end{tabular}
  }
\end{table}

\begin{table}[h! ]
  \caption{Comparison to SOTA on DrugComb - ZIP Synergy Score}
  \label{zip_table}
  \centering
  \resizebox{\textwidth}{!}{
  \begin{tabular}{lcccccc}
    \toprule
    \multirow{2}{*}{Method} & \multicolumn{2}{c}{Transductive} & \multicolumn{2}{c}{Leave-comb-out} & \multicolumn{2}{c}{Leave-drug-out}\\
    \cmidrule(l){2-7}
    & AUROC & AUPRC & AUROC & AUPRC & AUROC & AUPRC \\
    \midrule
    CongFu (ours) & $\mathbf{0.986 \pm 0.002}$ & $\mathbf{0.99 \pm 0.001}$ & $\mathbf{0.983 \pm 0.001}$ & $\mathbf{0.988 \pm 0.001}$ & $\mathbf{0.829 \pm 0.01}$ & $\mathbf{0.874 \pm 0.01}$ \\
    DeepDDS & $0.977 \pm 0.003$ & $0.983 \pm 0.002$ & $0.964 \pm 0.005$ & $0.974 \pm 0.004$ & $0.812 \pm 0.008$ & $0.86 \pm 0.01$\\
    DSN - DDI &  $0.947 \pm 0.01$ & $0.96 \pm 0.009$ & $0.964 \pm 0.002$ & $0.974 \pm 0.002$ & $0.793 \pm 0.024$ & $0.844 \pm 0.012$ \\
    XGBoost & $0.736 \pm 0.002$ & $0.742 \pm 0.001$ & $0.732 \pm 0.002$ & $0.739 \pm 0.005$ & $0.665 \pm 0.015$ & $0.689 \pm 0.027$ \\
    LogReg & $0.692 \pm 0.003$ & $0.712 \pm 0.002$ & $0.691 \pm 0.008$ & $0.711 \pm 0.007$ & $0.619 \pm 0.008$ & $0.66 \pm 0.025$ \\
    \bottomrule
  \end{tabular}
}
\end{table}

\newpage

\section{CongFu schematics}\label{app:schematics}

\subsection{CongFu-based model architecture}

\begin{figure}[h]
    \centering
    \includegraphics[width=1\textwidth]{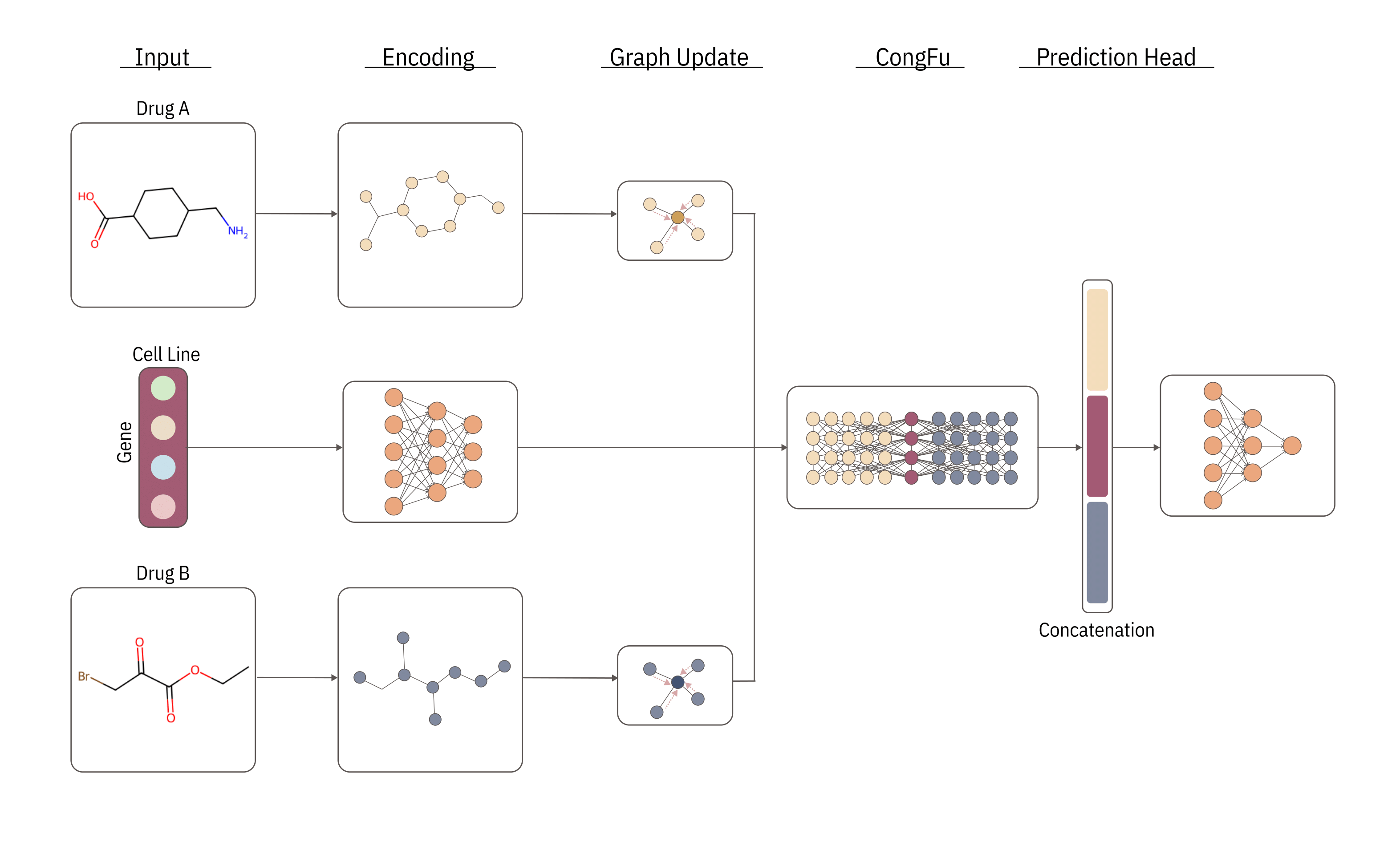}
    \caption{CongFu-based model architecture. The model takes in two drugs and a cell line as its input. The pairs of drugs are represented as graphs, where node features for each are obtained via atomic number embeddings, edge features are calculated as bond embeddings, and the cell line is encoded using MLP. Then, MPNN(s) are used to encode each graph separately. Next, CongFu layer(s) is utilized to fuse information from graphs and a cell line. Finally, an MLP is applied over the concatenation of drugs and the cell lines.}
    \label{congfu-model}
\end{figure}

\newpage

\subsection{CongFu-based model algorithm}

\RestyleAlgo{ruled}
\SetKwInput{KwInput}{Input}
\SetKwInput{KwOutput}{Output}

\begin{algorithm}[h]
\caption{Algorithm for $L$ layer CongFu-based model.}\label{alg:fusion}
\KwInput{Pair of graphs $\mathcal{G}_A$, $\mathcal{G}_B$ with $N_A$, $N_B$ nodes, and $\mathbf{E}_A$, $\mathbf{E}_B$ edges;
Adjacency matrices $\mathbf{A}_A \in \mathbb{R}^{N_A \times N_A}$ and $\mathbf{A}_B \in \mathbb{R}^{N_B \times N_B}$; Node features $\mathbf{X}_A \in \mathbb{R}^{N_A \times 1}$, $\mathbf{X}_B \in \mathbb{R}^{N_B \times 1}$, and edge features $\mathbf{E}_A \in \mathbb{R}^{E_A \times 1}$, $\mathbf{E}_B \in \mathbb{R}^{E_B \times 1}$; Context $\mathbf{C} \in \mathbb{R}^{1 \times D_{cont}}$; Layer $l \in [0,L-1]$; Fusion layer $F_l$.}
\medskip
\KwOutput{Node features $\mathbf{X}_A \in \mathbb{R}^{N_A \times D}$, $\mathbf{X}_B \in \mathbb{R}^{N_B \times D}$; Context $\mathbf{C} \in \mathbb{R}^{1 \times D}$;}
\medskip
$\mathbf{C} \gets \texttt{MLP}(\mathbf{C}) \in \mathbb{R}^{1 \times D}$\\
\smallskip
\For{$j \in [A,B]$}{
$\mathbf{X}_j \gets \texttt{NodeEncoder}(\mathbf{X}_j) \in \mathbb{R}^{N_j \times D}$\\
\smallskip
\For{$l=0,1,...,F_l-1$}{
    $\mathbf{E}_j \gets \texttt{EdgeEncoder}(\mathbf{E}_j) \in \mathbb{R}^{E_j \times D}$\\
    \smallskip
    $\mathbf{X}_{j} \gets \texttt{MPNN}_{l}(\mathbf{X}_{j}, \mathbf{E}_j, \mathbf{A}_j)$\\
    \smallskip
    $\mathbf{X}_{j} = \texttt{BatchNorm}(\mathbf{X}_{j})$\\
    \smallskip
    $\mathbf{X}_{j} = \texttt{ReLU}(\mathbf{X}_{j})$\\
    \smallskip
    }
}
\medskip
\For{$l=F_l,F_{l}+1,...,L-1$}{
\For{$j \in [A,B]$}{
$\mathbf{X}_{j} \gets \mathbf{X}_{j} + \mathbf{W}_1^l \mathbf{X}_{j} + \mathbf{W}_2^l \texttt{ReLU}(\mathbf{W}_3^l \mathbf{C})$\\
$\mathbf{X}_{j} \gets \texttt{MPNN}_{l}(\mathbf{X}_{j}, \mathbf{E}_j, \mathbf{A}_{j})$\\
$\mathbf{X}_{j} \gets \texttt{BatchNorm}(\mathbf{X}_{j})$\\
$\mathbf{X}_{j} \gets \texttt{ReLU}(\mathbf{X}_{j})$\\

\For{$i=0,1,...,N_{j}$}{
$\alpha_{i} \gets \texttt{softmax}(\mathbf{d}^{T}_l [\mathbf{W}_4^l \mathbf{C} || \mathbf{W}_4^l \mathbf{X}_{j}^{i}])$}
$\mathbf{C}_{j} \gets   \sum_{i}^{N_{j}}  \frac{1}{N_{j} + 1} \alpha_{i} \mathbf{W}_4^l \mathbf{X}_{j}^{i} + \mathbf{b}_l$\\
}
$\mathbf{C} \gets \mathbf{C}_A + \mathbf{C}_B$
}

\medskip
\Return $\texttt{MLP}([\mathbf{X}_A \big\| \mathbf{X}_B \big\| \mathbf{C}]) \in \mathbb{R}$
\end{algorithm}

\end{document}